\pdfoutput=1
\documentclass{article}
\usepackage{scml2024}

\usepackage[numbers]{natbib}
\usepackage[utf8]{inputenc} 
\usepackage[T1]{fontenc}
\usepackage{graphicx}%
\usepackage[utf8]{inputenc}
\usepackage{mathtools}
\usepackage{hyperref, hypernat}
\usepackage[todos]{CMImacros}
\usepackage{pgf}
\usepackage{wrapfig}
\usepackage[boxed]{algorithm2e} 
\usepackage{dsfont}
\graphicspath{{./figures/}} 

\newcommand\isPGF{F}
\newcommand\defaultSize{\if\isPGF T 0.5\else 0.49\fi}

\newcommand{\PGForPDF}[2][\defaultSize]{\if\isPGF T {\scalebox{#1}{\input{#2.pgf}}}\else {\includegraphics[width=#1\textwidth]{#2.pdf}}\fi}

\SetAlCapSkip{\abovecaptionskip}

\title{Learning phase-space flows using time-discrete implicit Runge-Kutta PINNs }%

\usepackage{authblk}

\author[1,2]{Álvaro Fernández Corral\thanks{Corresponding author: \texttt{alvaro.fernandez@cfel.de}}}%
\author[1]{Nicolás Mendoza}%
\author[3]{Armin Iske}%
\author[1,4]{Andrey Yachmenev}%
\author[1,2,4]{Jochen Küpper}%

\affil[1]{Center for Free-Electron Laser Science CFEL, Deutsches Elektronen-Synchrotron DESY,
	Notkestraße 85,
	22607
	Hamburg,
	Germany}

\affil[2]{Department of Physics, Universität Hamburg,
	Luruper Chaussee 149,
	22761
	Hamburg,
	Germany}

\affil[3]{Department of Mathematics, Universität Hamburg,
	Bundesstraße 55,
	20146
	Hamburg,
	Germany}

\affil[4]{Center for Ultrafast Imaging, Universität Hamburg,
	Luruper Chaussee 149,
	22761
	Hamburg,
	Germany}

\hypersetup{
	pdftitle={Learning phase-space flows using time-discrete implicit Runge-Kutta PINNs},
	pdfsubject={cs.AI, cs.LG, math.AP},
	pdfauthor={Álvaro Fernández Corral, Nicolás Mendoza, Armin Iske, Andrey Yachmenev and Jochen Küpper},
	pdfkeywords={Physics-informed neural networks, time-discrete methods, implicit Runge-Kutta, phase-space flows}
}

\begin{document}
\maketitle
\begin{abstract}
   We present a computational framework for obtaining multidimensional phase-space solutions of
   systems of non-linear coupled differential equations, using high-order implicit Runge-Kutta
   Physics-Informed Neural Networks (IRK-PINNs) schemes. Building upon foundational work originally
   solving differential equations for fields depending on coordinates [\emph{J.\ Comput.\ Phys.}
   \textbf{378}, 686 (2019)], we adapt the scheme to a context where the coordinates are treated
   as functions. This modification enables us to efficiently solve equations of motion for a
   particle in an external field. Our scheme is particularly useful for explicitly
   time-independent and periodic fields. We apply this approach to successfully solve the equations
   of motion for a mass particle placed in a central force field and a charged particle in a
   periodic electric field.
\end{abstract}

\keywords{physics-informed neural networks \and implicit Runge-Kutta methods \and phase-space flows}

\section{Introduction}
Physics-Informed Neural Networks (PINNs) have emerged as a prominent and dynamic area of research
for solving differential equations~\cite{Raissi:JCOP378:686, Cuomo:JSC92:88}, for example, for
modeling the physics of fluid dynamics~\cite{Jin:JCOP426:109951, Eivazi:PF34:075117, Rao:TAMC10:207}
or quantum mechanics~\cite{Shah:arXiv2210:12522}. Unlike traditional neural networks, that learn
solely from data, PINNs use both data and physical equations to guide the learning
process~\cite{Chen:NatComm12:6136, Xu:JCOP445:110592, Xu:PRR3:033270}.

PINNs can be effectively employed using continuous and discrete representations of time. The
time-continuous approach uses space and time variables as inputs, and learn to satisfy the
differential equations across the entire domain of interest. This can be impractical without data
distributed across multiple time slices. In addition, time-continuous PINNs also encounter
difficulties with high-frequency oscillations and stiff problems, lacking a clear strategy to deal
with them. On the other hand, the discrete-time PINNs learn to model changes within a fixed discrete
time step, utilizing only spatial information from a single time slice. This approach improves the
accuracy in solving stiff problems~\cite{Sharma:ArchCME:30:2929} by leveraging the $A$-stability of
implicit Runge-Kutta (IRK) methods~\cite{Hairer:StiffProblems}. This is especially significant for
tackling the stiff problems prevalent in particle trajectory simulations and Differential Algebraic
Equations (DAEs), known for their inherent infinite stiffness~\cite{Hairer:StiffProblems,
   Kim:Chaos31:093122}.

In this paper, we extend the discrete-time IRK-PINN scheme~\cite{Raissi:JCOP378:686} by generalizing
it to a larger number of dimensions in both input variables and output quantities. Importantly, we
have adapted this approach to develop a new efficient numerical scheme tailored to finding
phase-space flows of classical particle trajectories, described by second-order differential
equations of motions. The new algorithm enables the simultaneous learning of all possible
trajectories within the specified phase-space and is particularly suited for explicitly
time-independent and time-periodic forces.

The manuscript is organized as follows: The details and application of the IRK-PINN scheme to
phase-space flows are described in \autoref{sec:trajectory_finding}. \autoref{sec:results} presents
some illustrative application examples with time-independent and periodic forces. The review and
application details of the IRK-PINN scheme, including its successful application to a number of
various differential equations, are elaborated in
\autoref{app:runge_kutta_scheme}--\ref{app:additional_results}.

\section{Phase-space flows with IRK-PINNs}\label{sec:trajectory_finding}
In applications to physics problems, the IRK-PINN method was primarily applied to functions
$\mathbf{u}:\mathds{R}^{1+d}\to \mathds{R}^m$ which represent background ``fields'' of the system,
such as the temperature, pressure, velocity flow, heat convection, etc., see
\autoref{app:runge_kutta_scheme} for details. The literature on applying PINN-based IRK schemes to
particle trajectory analysis in physical systems is limited, with the exception of work focusing on
first-order differential equations~\cite{Moya:NCA35:3789}.
We are interested in solving differential equations of the general form
\begin{align}
    \label{eq:trajectory_finding_general}
    \mathbf{F}(\mathbf{x}, \dot{\mathbf{x}}, \ddot{\mathbf{x}},\ldots, t) = \mathbf{0},
\end{align}
which encompass, for example, Newton's equation, expressed as
$\ddot{\mathbf{x}} = \mathbf{f}(\mathbf{x}, \dot{\mathbf{x}}, t)$. Our goal is to adapt and apply
the IRK-PINNs scheme to effectively compute particle flows $\mathbf{x}(t)$ resulting from a force
$\mathbf{f}(\mathbf{x}, \dot{\mathbf{x}}, t)$. We begin by transforming the second-order Newton-type
differential equation into a system of first-order differential equations using the $2d$-dimensional
phase-space coordinates, denoted as $\mathbf{\chi} = (\mathbf{x}, \mathbf{\dot{x}})$, with the
dimensionality $d$ of the vector $\mathbf{x}$. We define our function of interest as
$\mathbf{u}(t, \mathbf{\mathbf{\chi}}_i) = \hat{\mathbf{\chi}}_i(t)$, with
$\hat{\mathbf{\chi}_i}:\mathds{R}\to \mathds{R}^{2d}$ representing the flow function. This flow
function defines the trajectory of any point in phase-space -- a one parameter curve in our
phase-space manifold -- that satisfies the condition
$\hat{\mathbf{\chi}}_i(t_n) = \mathbf{\mathbf{\chi}}_i$. By adhering to the equations of motions,
this condition ensures a unique solution. Indeed, since the value of $\mathbf{u}(t_n,\chi_i)$ at the
initial time $t_n$ is $\mathbf{u}_n(\mathbf{\chi}_i) = \mathbf{\chi}_i$, we can determine the
phase-space values at time $t_{n+1}$ by employing the IRK-PINNs time propagation method. By applying
the IRK scheme, see \eqref{eq:runge_kutta_scheme} in \autoref{app:runge_kutta_scheme}, using a fully connected
feedforward neural network (FNN) also known as a multilayer perceptron (MLP),  to the
general trajectory equation \eqref{eq:trajectory_finding_general}, we thus determine the
trajectories that satisfy the following set of differential equations
\begin{align}\label{eq:trajectory_finding_newton}
  0 = \mathcal{N}[\mathbf{\chi}]+\frac{d}{dt}\mathbf{\chi} \equiv \begin{pmatrix}\mathbf{\dot{x}}\\ \mathbf{f}(\mathbf{x},\mathbf{\dot{x}}, t) \end{pmatrix}+\frac{d}{dt} \begin{pmatrix}\mathbf{x}\\ \dot{\mathbf{x}} \end{pmatrix}.
\end{align}
Any additional boundary conditions that may be essential for the well-defined nature of the problem
can also be accommodated.

The algorithm's strength lies in its ability to predict the future state of every point in the
phase-space at time $t_{n+1}$ by learning from only a limited sample of phase-space data at the
initial time $t_n$. In addition, as a high-order IRK algorithm, it is well suited to solve stiff
problems, enhancing computational efficiency in the propagation of numerous trajectories.

Moreover, this algorithm is particularly effective for forces that are explicitly time-independent
and/or periodic in time. If the force doesn't explicitly change over time and only relies on the
position or velocity of the particle, both being implicitly time-dependent vector fields, the neural
network can effectively predict the trajectories by learning from a training set in the phase-space.
For periodic forces, we impose the time-propagation step size to match the period of the force
$\Delta t \coloneqq T$, so that the neural network output $\chi(t_{n+1})$ would reside on a
phase-space manifold identical to the initial one at $t_{n}$. The periodic nature of the force is
essential, as the phase-space manifold is entirely dependent on the force governing the particles'
motions, necessitating that $\mathbf{f}(\mathbf{x}, t_{n}+\Delta t)$ remains equivalent to
$\mathbf{f}(\mathbf{x}, t_{n})$. In both cases, or a combination, the phase-space manifold remains
unchanged after a single time step propagation. Consequently, to predict trajectories for subsequent
time slices, the IRK-PINN algorithm can be recurrently applied after training.

\section{Results}
\label{sec:results}
\subsection{Keplerian orbits}
A massive body under the action of a central potential moves according to Newton's equation of
gravity. The trajectories that orbiting objects follow are called Keplerian orbits. The analytical
solution of $N$-body problems is complicated due to the implicit time-dependency of the force
$F\sim 1/r_{ij}^2(t)$, arising from the varying distances between different objects $r_{ij}^2(t)$. For this reason, low-order IRK schemes with small time stepping were used to propagate these systems~\cite{Antonana:CMDA134:3}.
Following the discussion in the previous section, our phase-space flows approach offers an efficient
alternative for propagating these explicitly time-independent systems. This is achieved by
recurrently applying the neural network, requiring only a single training session of the IRK-PINN
and thus eliminating the need for additional approximations.

We focus on central forces, which conserve angular momentum, leading to the confinement of
trajectories within a two-dimensional plane defined by the initial momentum and the radial
vector~\cite{Goldstein:ClassicalMechanics}. Initially, our efforts were directed at solving the
fixed-central-mass problem in a two-dimensional Euclidian space with a Coulomb potential. However,
after experimenting with various MLPs and parameters, we
observed that the algorithm was unable to accurately predict the trajectories for this problem. We
believe that this limitation arises from the fact that not all points in the initial phase-space
correspond to physically viable solutions. Given that the algorithm attempts to simultaneously solve
for all trajectories, these divergent paths likely cause a general diverging behaviour of the
method.

\begin{figure}
	\centering%
	\includegraphics[width=\linewidth]{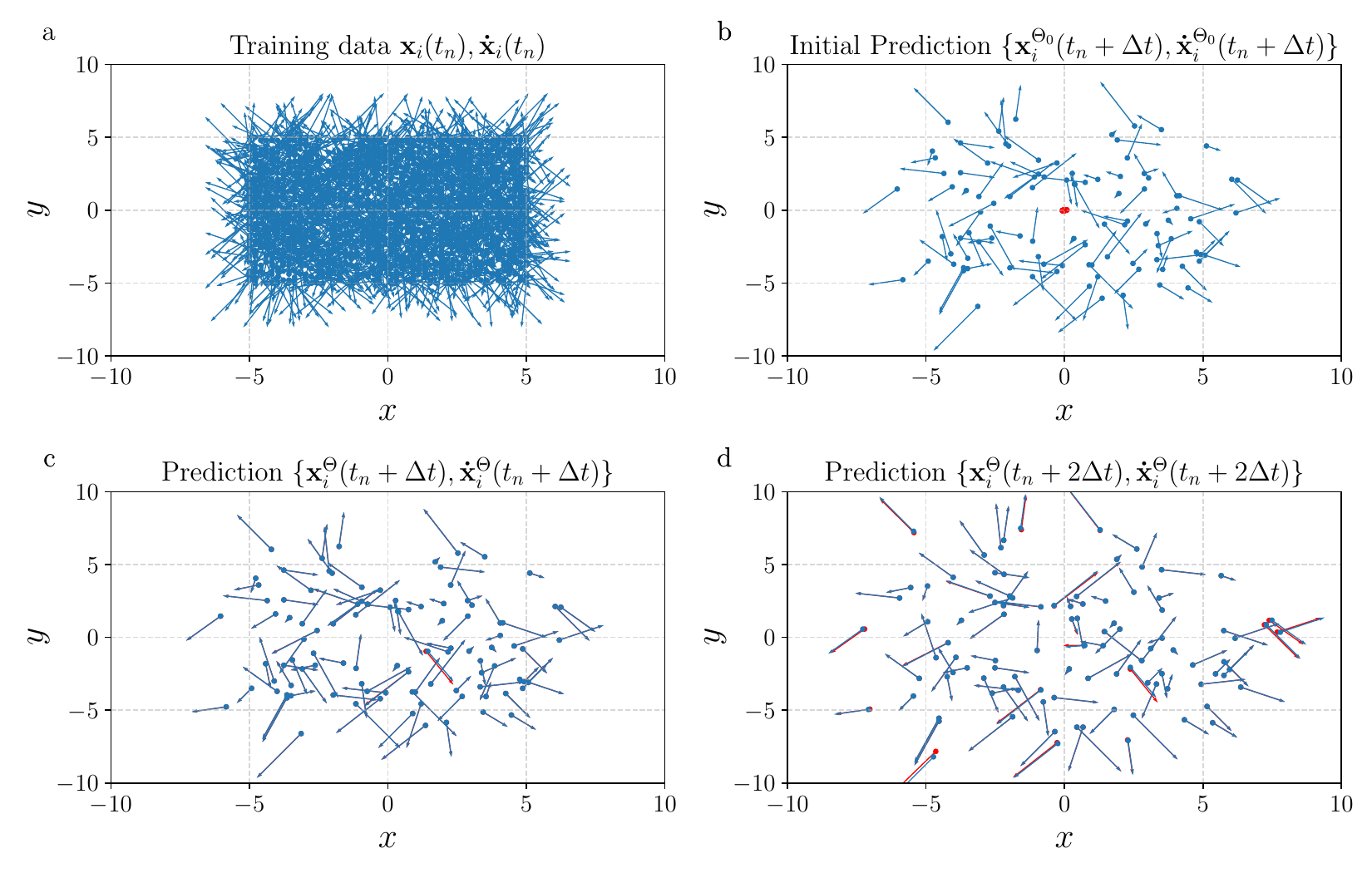}
	\caption{Calculated trajectories of a particle in a central Gaussian potential, with potential
		depth $V_0 = 10$ and extension $a = \sqrt{5}$. The IRK-PINN and numerically exact solutions
		are plotted with blue and red colours, respectively. Panel (a) shows the entire training phase
		space consisting of 2\,000 points. Due to the complex distribution of the solutions with
		respect to the initial position, a big set of training data is needed. Panel (b) shows the
		result of the low-order IRK method (blue) on the validation set, consisting of 100 points, for
		visual clarity. In addition, the PINN guess on parameter initialisation is displayed (red).
		Panels (c) and (d) show the IRK-PINN predictions on the validation set for the time step
		$\Delta t$ and $2 \Delta t$, respectively. The IRK-PINNs approach adopted a Runge-Kutta order
		$q=100$, time step $\Delta t = 0.8$, dense MLP containing 5 hidden layers with 200 nodes each,
		and a bipolar sigmoid as the activation function. The training was carried out for 100\,000
		epochs with the ADAM optimiser followed by 40\,000 epochs with L-BFGS-B. Total
		$L^1\text{-error}=0.58~\%$.}
	\label{fig:Gauss_kepler}
	\vspace{-3pt}
\end{figure}

The accurate and efficient calculation of Keplerian orbits is critical in simulations of particle
diffraction from the Coulomb potential with a small impact parameter. This is especially important
for simulations of laser-induced electron diffraction~\cite{Blaga:Nature483:194}, where the
diffraction of highest-energy electrons is influenced by these orbital
dynamics~\cite{Shvetsov-Shilovski:EPJD75:130, Wiese:PRR3:013089, Wiese:thesis:2020}.

To avoid singularities, we replaced the Coulomb potential with a central potential characterized by
a Gaussian function $V (\mathbf{x}) = - V_0 \exp(-\mathbf{x}^2/a^2)$, where $V_0$ is the depth of the potential well and the $a$ parameter
determines its extent. This choice of potential is a standard first-order approximation to the Coulomb potential in strong-field physics~\cite{Toulouse:IJQC100:10471056}
and provides Kepler-like orbits, while not provoking diverging
trajectories, due to its finite value at the origin. Defining the force as the negative gradient of
this potential, the equations of motion become
\begin{align}\label{eq:gaussian_kepler_diff_eq}
	\mathcal{N}[\mathbf{\chi}] &= -\begin{pmatrix} \dot{\mathbf{x}}\\ -\nabla_{\mathbf{x}}V(\mathbf{x}) \end{pmatrix} \equiv -\begin{pmatrix} \dot{\mathbf{x}}\\ - 2 \frac{V_0}{a^2}  \exp(-\frac{ \mathbf{x} ^2}{a^2} )  \mathbf{x}  \end{pmatrix}.
\end{align}

The domain of the model is
$(\mathbf{x}, \dot{\mathbf{x}})\in\Omega = \mathds{R}^2\times \mathds{R}^2$. Considering that the
phase-space flows approach uses an equal number of phase-space coordinates and objective functions,
we will work within a 4-dimensional phase-space and output space. For benchmarking the IRK-PINN
results, we generated accurate solutions using low-order adaptive IRK schemes, specifically
Kv{\ae}rn{\o}'s 5/4 method~\cite{Kvaerno:BITNM44:489}, as implemented in the Diffrax
library~\cite{Kidger:thesis:2021}.

The accuracy of the IRK-PINNs solution, along with the details of the neural network, are
demonstrated in~\autoref{fig:Gauss_kepler}. It is important to note that the central force in this
model is explicitly time-independent. This aspect, combined with comprehensive solution for the
entire phase space, enables us to propagate in time steps of $\Delta t$ by recurrently applying the
model at each time step. To illustrate this, in~\autoref[d]{fig:Gauss_kepler} we show the accuracy
of predictions of the neural network at a doubled time step $2\Delta t$. In addition, this panel
shows the ability of the PINN to predict trajectories from points that are not contained inside the
training region.

\subsection{Charged particle under the action of a sinusoidal laser}

In this example, we consider the motion of a charged particle under the influence of a periodic external
electric field, \ie, a laser field. The force exerted on a particle with charge $q$ by an external field
$\mathbf{E}(\mathbf{x},t)$ is given by $\mathbf{F}(\mathbf{x},t) = q\mathbf{E}(\mathbf{x},t)$. As in
the previous example, we continue to work within a $4$-dimensional phase space defined by
$(\mathbf{x}, \dot{\mathbf{x}})\in\Omega = \mathds{R}^2\times \mathds{R}^2$, also corresponding to a
4-dimensional output space.

\begin{wrapfigure}{r}{0.5\textwidth}
	\includegraphics[width=0.5\textwidth]{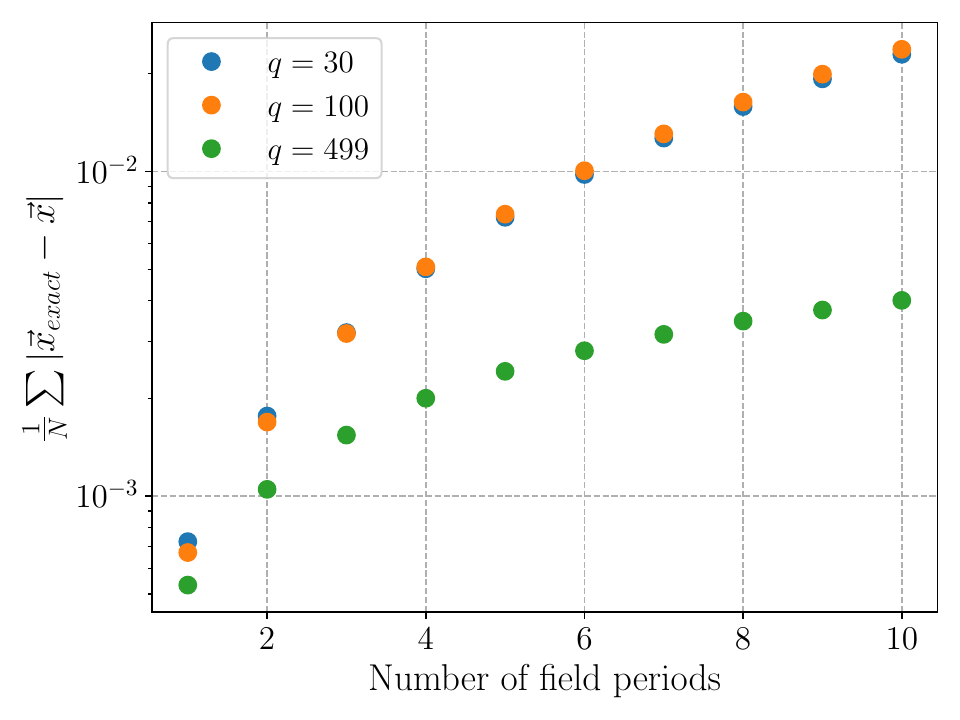}
	\caption{Accuracy of IRK-PINNs method for different Runge-Kutta orders $q=30,100,499$ in solving
		the trajectories of a charged particle in periodic electric field, as described by
		\eqref{eq:charged_particle_diff_eq} with the parameters $\alpha=0.5$ and $A=10$. Accuracy is
		quantified using the $L^1$-error calculated on a validation set of 2\,500 points. Training
		involved 50\,000 epochs each with the ADAM and L-FBGS-B optimizers. The model architecture
		included an MLP with 5 hidden layers, each composing 32 nodes, and employed the $SiLU(x)$
		activation function. The training utilized 5\,000 points.}
	\label{fig:Periodic_force}
\end{wrapfigure}

For simplicity, we assume that the wavelength of the laser field is much larger than the scale of particle
movement. The electric field, characterised by an angular frequency $\omega$ and incident at an
angle $\alpha$ relative to the $x$-axis, is represented by the function
$\mathbf{E}(\mathbf{x}, t) =( E_x(t), E_y(t)) =( A \cos(\omega t) \cos(\alpha), A \cos(\omega t)
\sin(\alpha))$, where $A$ denotes field's amplitude. Choosing the unit system $M = q = 1$, the
differential equation of motion can be expressed as
\begin{align}
   \mathcal{N}[\mathbf{\chi}] = -\begin{pmatrix} \dot{\mathbf{x}}\\ \mathbf{E}(\mathbf{x}) \end{pmatrix} = -\begin{pmatrix}\dot{x}\\\dot{y}\\ A \cos(\omega t) \cos(\alpha)\\ A \cos(\omega t) \sin(\alpha) \end{pmatrix}.
   \label{eq:charged_particle_diff_eq}
\end{align}

This particular form of the laser field is chosen for its analytical solvability, substantive
complexity, and periodic behaviour in time. Given any initial conditions
$(x_0, y_0, \dot{x_0}, \dot{y_0})$ at time $t_0$, the analytical solution can be readily obtained.
We examine the system with a laser of period $T =1 \coloneqq \Delta t$, amplitude $A = 10$ and
incident angle $\alpha = 0.5$. To benchmark our IRK-PINNs implementation, we compare its results for
different Runge-Kutta orders with the analytical solution. The results of the simulations and the
details of the neural networks used are presented in \autoref{fig:Periodic_force}. These results
were obtained for multiple periods of the electric field by recurrently applying the neural network,
which was trained only once, for the first period, using a set of training points distributed across
a subset of the phase-space. For the purpose of illustration and avoiding overfitting, a validation
set contained in a different subset of the phase-space was used to generate the plots.

\section{Conclusions}
We introduced a versatile algorithm designed to effectively solve a broad range of differential
equations. The algorithm was validated by generating accurate results for both functional PDEs and equations of motion. Notably, the application of PINNs as a propagator
for explicitly time-independent and periodic forces represents a significant advancement over
conventional low-order IRK methods.

Further work should focus on addressing the problem of divergent trajectories, particularly in cases
like Keplerian orbits under a Coulomb $\sim1/\abs{\mathbf{x}}$ potential. Overcoming this
divergences would particularly enhance the algorithm's utility in solving stiff dynamical systems,
such as the $N$-body problem~\cite{Antonana:CMDA134:3} and the dynamics of charged particles in time-independent or periodic
external fields~\cite{Shvetsov-Shilovski:EPJD75:130, Wiese:PRR3:013089}.

\clearpage

\section*{Acknowledgements}
This work was supported by Deutsches Elektronen-Synchtrotron DESY, a member of the Helmholtz
Association (HGF), including the Maxwell computational resource operated at DESY, by the Data
Science in Hamburg HELMHOLTZ Graduate School for the Structure of Matter (DASHH, HIDSS-0002), and by
the Deutsche Forschungsgemeinschaft (DFG) through the cluster of excellence ``Advanced Imaging of
Matter'' (AIM, EXC~2056, ID~390715994).

\bibliographystyle{apsrev4-2}
\bibliography{string,cmi}

\clearpage

\appendix
\section{High-order implicit Runge-Kutta scheme}
\label{app:runge_kutta_scheme}
We start by examining a vector-valued function $\mathbf{u}: \Sigma \rightarrow \mathds{R}^m$, which
is defined over spacetime vectors
$(t, \mathbf{x}) \in \Sigma \coloneqq \mathds{R} \times \Omega \subset \mathds{R}^{1+d}$ in the
domain formed by a combination of $1$-dimensional time and $d$-dimensional space. This function is
defined to be the solution of a set of non-linear coupled differential equations, represented as
$\partial_t\mathbf{u} + \mathcal{N}[\mathbf{u}] = 0$, $\forall (t,\mathbf{x})$ in $\Omega$.
Additionally, we considered the possibility of incorporating boundary conditions, denoted as
$\mathcal{B}_\alpha[\mathbf{u}(\partial \Omega)] = 0$, which could depend on the derivatives of
$\mathbf{u}$ at the boundary.

We sought to develop an algorithm that utilizes the dataset
${\{\mathbf{x}_k, \mathbf{u}(t_n, \mathbf{x}_k)\}_{k=1}^N}$ at a specific time slice $t_n$ and the
differential equations representing the time evolution to accurately predict
$\mathbf{u}(t_{n+1}, \mathbf{x})$ at the next time slice $t_{n+1}$. To describe the time propagation
from $t_n$ to $t_{n+1}$, we introduced an IRK scheme of $q$-th order, defined by a set of coupled
equations
\begin{subequations}\label{eq:runge_kutta_scheme}
	\begin{align}
		\label{eq:runge_kutta_scheme_a}
		\mathbf{u}_{n+c_i} &= \mathbf{u}_n -  \Delta t \sum_{j=1}^q a_{ij} \mathcal{N}[\mathbf{u}_{n+c_j}] \quad \forall i \in \{1, \ldots, q\}, \\
		\label{eq:runge_kutta_scheme_b}
		\mathbf{u}_{n+1} &= \mathbf{u}_n - \Delta t \sum_{j=1}^q b_j \mathcal{N}[\mathbf{u}_{n+c_j}],
	\end{align}
\end{subequations}
where $\mathbf{u}_{n+c_i}(\mathbf{x})\coloneqq \mathbf{u}(t_n+c_i\Delta t, \mathbf{x})$,
$\Delta t=t_{n+1}-t_n$ and $a_{ij}$, $b_j$, and $c_{i}$ are the Butcher-tableau coefficients for a
chosen IRK order $q$. The computation of the Butcher tableau involves expanding the solution into a
Taylor series and matching the coefficients to the actual solution up to the desired order of
accuracy. The choice of this coefficients is not unique, and it specifies the particular IRK method.
We chose the Gauss-Legendre Runge-Kutta method, which is $A$-stable for all
orders~\cite{Iserles:FirstCourseNumAnalDiffEq}, to find the coefficients throughout our
implementation. However, other methods, such as the Lobatto~\cite{Gonzalez:JCAM82:129148}, the
Radau~\cite{Hairer:JCAM111:93111}, or the diagonally implicit Runge-Kutta
methods~\cite{Kennedy:ANM146:221244}, result in different accuracy, stability, and efficiency
properties. Testing the pros and cons of using different Butcher-tableau coefficients in
applications with IRK-PINNs is beyond the scope of this manuscript, although it should be further
explored in the future.

The theoretical analysis of the IRK algorithm suggests that the deviation from the exact result
scales as $\mathcal{O} (\Delta t ^{2 q})$~\cite{Iserles:FirstCourseNumAnalDiffEq}. Although it may
initially appear that the error would grow with increasing $q$ for time steps $\Delta t>1$, it is
important to recognize that this error is dimensionful and requires a constant to render it
dimensionless for proper interpretation.

For a chosen IRK order $q$, we place a neural network prior on all intermediate calculations and the
final output
\begin{align}
	\mathcal{U} \coloneqq (\mathbf{u}_{n+c_1}, \ldots, \mathbf{u}_{n+c_q}, \mathbf{u}_{n+1} ).
	\label{eq:NN_prior}
\end{align}
We observe that $\mathcal{U}(\mathbf{x}) \in \mathds{R}^{m\times ( q+1)}$ can be conceptualized as
an $m \times (q+1)$ matrix, where $m$ is the output dimension of $\mathbf{u}$ which is not
necessarily equal $1$. Consequently, we employ an MLP denoted by
$\mathcal{U}^\theta:\mathds{R}^d \to \mathds{R}^{m\times( q+1)}$, to closely approximate the desired
function $\mathcal{U}$. We define $\mathbf{u}^\theta_j$ in a manner analogous to \eqref{eq:NN_prior}
as the parameter-dependent approximation. Subsequently, we define a set of parameter-dependent
quantities to be used in the loss function as
\begin{subequations}\label{eq:RK_approximants}
	\begin{align}
		\label{eq:RK_approximants_a}
		&\mathbf{k}^\theta_i = \mathbf{u}^\theta_{n+c_i}+\Delta t \sum_{j=1}^q a_{ij} \mathcal{N}[\mathbf{u}^\theta_{n+c_j}] \quad \forall i \in \{1, \ldots, q\} \\
		\label{eq:RK_approximants_b}
		&\mathbf{k}^\theta_{q+1} = \mathbf{u}^\theta_{n+1} + \Delta t \sum_{j=1}^q b_j \mathcal{N}[\mathbf{u}^\theta_{n+c_j}].
	\end{align}
\end{subequations}

Comparing \eqref{eq:RK_approximants} with \eqref{eq:runge_kutta_scheme}, it is evident that if
$\mathcal{U}^\theta$ effectively approximates $\mathcal{U}$, then the earlier definition implies
${\mathbf{k}^\theta_{i} \approx \mathbf{u}_{n}}\;\forall i\in\{1, \ldots, q+1\}$. Using this
relationship, we can express the loss function as
\begin{align}
	\mathcal{L}_{\mathcal{N}}(\theta, \{\mathbf{x}_k\}_{k=1}^N) = \frac{1}{N}\sum_{k=1}^N \sum_{i=1}^{q+1} \left\| \mathbf{k}^\theta_{i}(\mathbf{x}_k) - \mathbf{u}_{n}(\mathbf{x}_k) \right\|^2,
	\label{eq:loss_function_diff_eq}
\end{align}
where $\{\mathbf{x}_k\}_{k=1}^N$ is a set of spatial points randomly distributed in the phase-space
$\Omega$.

Furthermore, if the differential equations include $A$ boundary conditions of the type
$\mathcal{B}_\alpha[\mathbf{u}(\partial \Omega)] = 0$, an additional term can be added to the loss
function
\begin{align}
	\mathcal{L}_\mathcal{B}(\theta,   \{\mathbf{\hat{x}}_b\}_{b=1}^B) = \frac{1}{B} \frac{1}{A} \sum_{b=1}^B \sum_{\alpha=1}^A \left\| \mathcal{B}_\alpha[\mathbf{u}(\mathbf{\hat{x}}_b)] \right\|^2,
	\label{eq:loss_function_boundary}
\end{align}
where $\{\mathbf{\hat{x}}_b\}_{b=1}^B$ is a set of spatial points distributed in the boundary
$\partial \Omega$.

The total loss function used in the optimization is expressed as a weighted sum of the
aforementioned terms, as
\begin{align}
	\mathcal{L}(\theta, \{\mathbf{x}_k\}_{k=1}^N, \{\mathbf{\hat{x}}_b\}_{b=1}^B) = \omega_{\mathcal{N}} \mathcal{L}_{\mathcal{N}} + \omega_\mathcal{B}	\mathcal{L}_\mathcal{B}.
	\label{eq:loss_function_total}
\end{align}
To ensure best convergence, the ratio of the weights $\omega_\mathcal{N}$ and $\omega_\mathcal{B}$
can be chosen different depending on the problem.

\section{Methodology}
\label{app:methodology}
We implemented the IRK-PINNs using the JAX~\cite{jax2018github} and Flax~\cite{Flax:Software2020}
Python libraries and optimized the weights of the PINN by initially approaching the minimum using
Optax~\cite{Optax:DeepMindJax} with the Adam first-order optimizer~\cite{Kingma:ICLR2015} and then
refined the result by switching to the second-order optimization L-BFGS-B~\cite{Byrd:SISC16:1190}
available in the Jaxopt library~\cite{Jaxopt:Software2022}. For the PINN, we employed a
fully-connected dense MLP with different number and structure of hidden layers and types of
activation functions, depending on the problem addressed. We found that
$SiLU(x)= \frac{x \exp(x)}{1+\exp(x)}$ and bipolar sigmoid
${f(x) = \frac{\exp(x)-1}{\exp(x)+1}}$~\cite{Panicker:IOSRJE2:1352} activation functions work best
for the problems considered in this study.

For benchmarking, we obtained accurate results using the low-order Runge-Kutta approach from the
Diffrax library~\cite{Kidger:thesis:2021}, which provides various implicit and explicit Runge-Kutta
methods of different orders.

Our developed model is highly versatile and easy to use. Using just one class type, it successfully
handled a range of applications, some of which are highlighted in~\autoref{app:additional_results}.

\section{Additional results}
\label{app:additional_results}
In this section, we present the performance of the IRK-PINNs scheme applied to systems that handle
fields $\mathbf{u}:\mathds{R}^{1+d}\to \mathds{R}^m$, similar to those investigated
previously~\cite{Raissi:JCOP378:686}. These differ conceptually from a particle's equations of
motion, which deal with coordinates
$(\mathbf{x}, \mathbf{\dot{x}}): \mathds{R} \to \mathds{R}^{2d}$, in the manner that fields
do not have to follow a direct relation with the phase-space of the
system.

In the following, we selected a number of functional PDEs with different input/output dimensions to
illustrate the versatility of the algorithm.

\subsection{Heat Equation in a 2D Plate}
We deal with a system with a $2$-dimensional input: the heat equation over a surface. This equation
describes the evolution of a scalar field $T : \mathds{R}^{2+1}\to \mathds{R}$, representing the
temperature of a 2D system in time. Using our formalism, the heat equation is given by
\begin{align}
	\mathcal{N}[T] = -c^2\nabla^2 T,
	\label{eq:heat_eq}
\end{align}
where $c$ is the thermal diffusivity, a constant that measures the rate of heat transfer inside the
material.

We will deal with the system shown in~\cite{Daileda:SlidesHeat}, which is a two dimensional sheet
$(x, y)\in\Omega ={[0, 2]}^2 $ with $c=1/3$ and periodic boundary conditions for the temperature
given by $T(t, 0, y) = T(t, 2, y)$ = $T(t, x, 0) = T(t, x, 2)=0$. We will also impose the initial
condition of a heated lower half plane $ T(0, x, y) = 50(1-\Theta(y-1))$, with $\Theta$ being the
unit (Heaviside) step function. In essence, this represents a $2\times 2$ square that is initially
heated and in contact with a cold resevoir at its boundary with $T=0$. The solution to this equation
is given by
\begin{align}
	\begin{split}
		T(t, x, y) = \sum_{m, n=1}^\infty & \frac{200}{\pi^2} \frac{(1+(-1)^{m+1})(1-\cos\frac{n\pi}{2})}{mn}\cdot\\
		&\cdot\sin \frac{m\pi x}{2} \sin\frac{n\pi y}{2}e^{-\frac{c^2\pi^2}{4}(m^2+n^2)t}.
	\end{split}\label{eq:heat_eq_solution}
\end{align}

\begin{figure}
   \includegraphics[width=\linewidth]{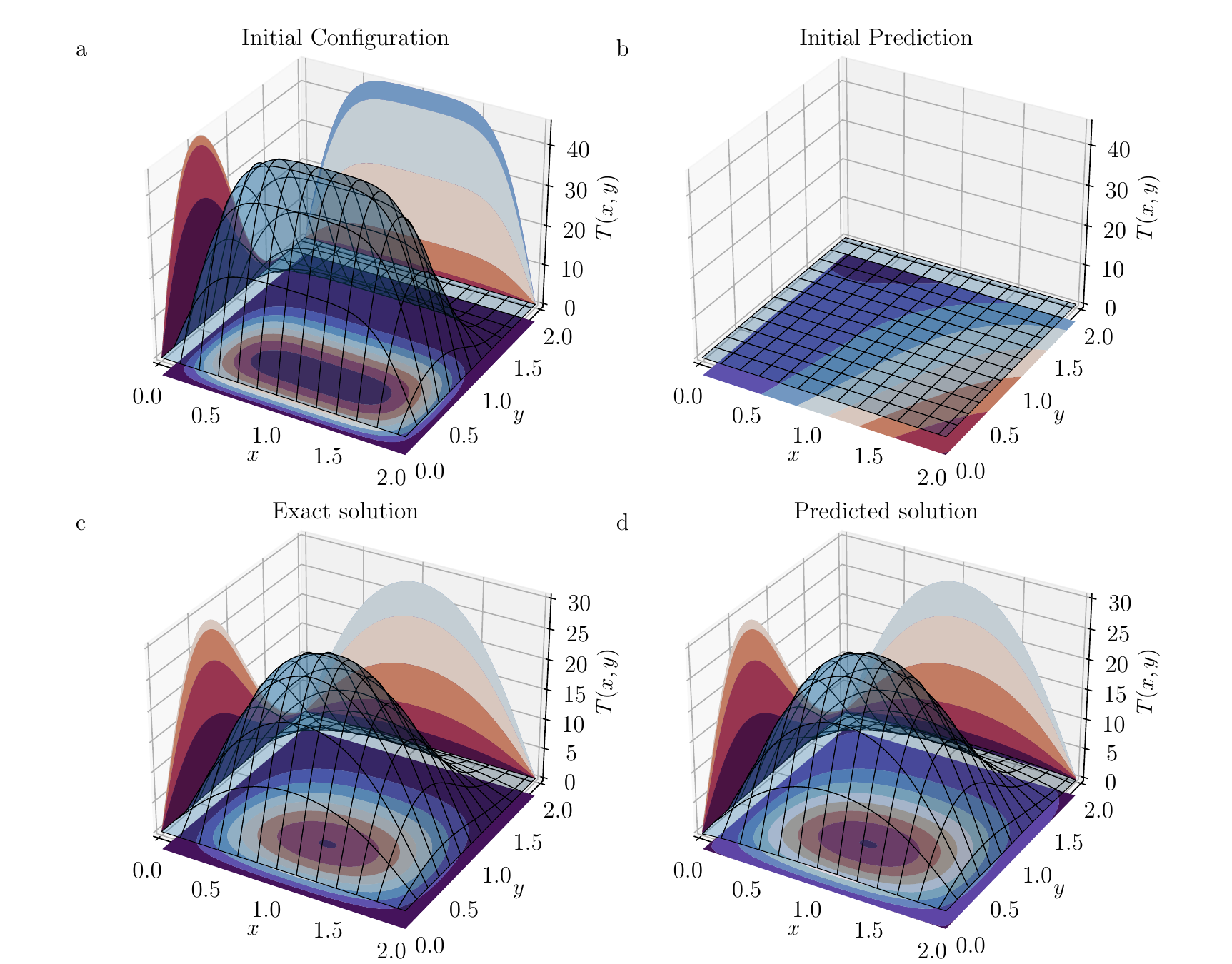}
   \caption[short]{Fit to the heat equation data with IRK-PINNs of order $q=100$ and time step
      $\Delta t = 0.7$, compared to the analytical solution of the problem. The training was
      performed for 25000 epochs with ADAM, 40000 epochs with L-BFGS-B, using an MLP with 20 hidden
      layers with 32 nodes each and $SiLU$ as activation function. (a) The initial configuration
      space, consisting of $N=500$ points. (b) The predition of the PINN on parameter
      inisialization. This initial prediction is totally arbitrary and far from the correct result,
      which shows the robustness of the learning process. (c) The analytical solution of the heat
      equation after one time step. (d) The prediction of the IRK-PINN after training. Total $L^1$
      error $=4.81\%$}
   \label{fig:heat_eq}
\end{figure}
The results of our IRK-PINN prediction, compared to~\eqref{eq:heat_eq_solution} and the description
of the used neural network, are shown in \autoref{fig:heat_eq}. The IRK-PINN is able to learn the
solution of the heat equation, even for relatively small neural network and sample size.

\subsection{Incompressible Navier-Stokes equation: Taylor-Green vortices}
The Navier-Stokes Equations describe the motion of Newtonian fluids. In these equations, the
variation of the quantity of fluid and its velocity are studied, usually in a compact or periodic
domain. The pressure, temperature and density of the fluid also play a role in the description, as
well as its viscosity. The search for general solutions to this set of equations is still a very
active field of research, with analytical solutions being rarely available for specific assumptions.
We study the Taylor-Green vortex~\cite{Taylor:PRSA158:499}, which is an incompressible Navier-Stokes
equation that has an exact closed form in Cartesian coordinates. The Taylor-Green vortex system is
described by
\begin{subequations}\label{eq:Taylor_Green_diff_eqs}
	\begin{align}
		\label{eq:Taylor_Green_diff_eqs_a}
		\mathcal{N}[u] &=  u \partial_x u + v \partial_y u + \frac{1}{\rho} \partial_x p - \nu (\partial_x ^2 u + \partial_y^2 u )\\
		\label{eq:Taylor_Green_diff_eqs_b}
		\mathcal{N}[v] &=  u \partial_x v + v \partial_y v + \frac{1}{\rho} \partial_y p - \nu (\partial_x ^2 v+ \partial_y^2 v ),
	\end{align}
\end{subequations}
where $u(t, x, y)$ and $v(t, x, y)$ are respectively the $x$ and $y$ components of the velocity
fields of the fluid, $ \nu $ is the viscosity, $ \rho $ is the mass density, and $ p(t, x, y)$ is
the pressure of the fluid. To account for the continuity equation for these velocities, which is
$\partial_x u + \partial_y v = 0$, we treat it as a boundary condition and add it to our algorithm
as an extra term to the loss function.

By restricting the domain of the solutions to the 2-dimensional plane
$\mathbf{x}\in\Omega = {[0, 2 \pi]}^2$, the analytical solution takes the form
\begin{subequations}
	\begin{align}
		u(t, x, y) &= \sin(x) \cos(y) \exp(-2 \nu t)\\
		v(t, x, y) & = -\cos(x) \sin(y) \exp(-2 \nu t)\\
		p(t, x, y) &= - \frac{\rho}{4} (\cos 2x  + \cos 2y  ) \exp(-4 \nu t).
	\end{align}
	\label{eq:TaylorVortex}
\end{subequations}

\begin{figure}
   \includegraphics[width=\linewidth]{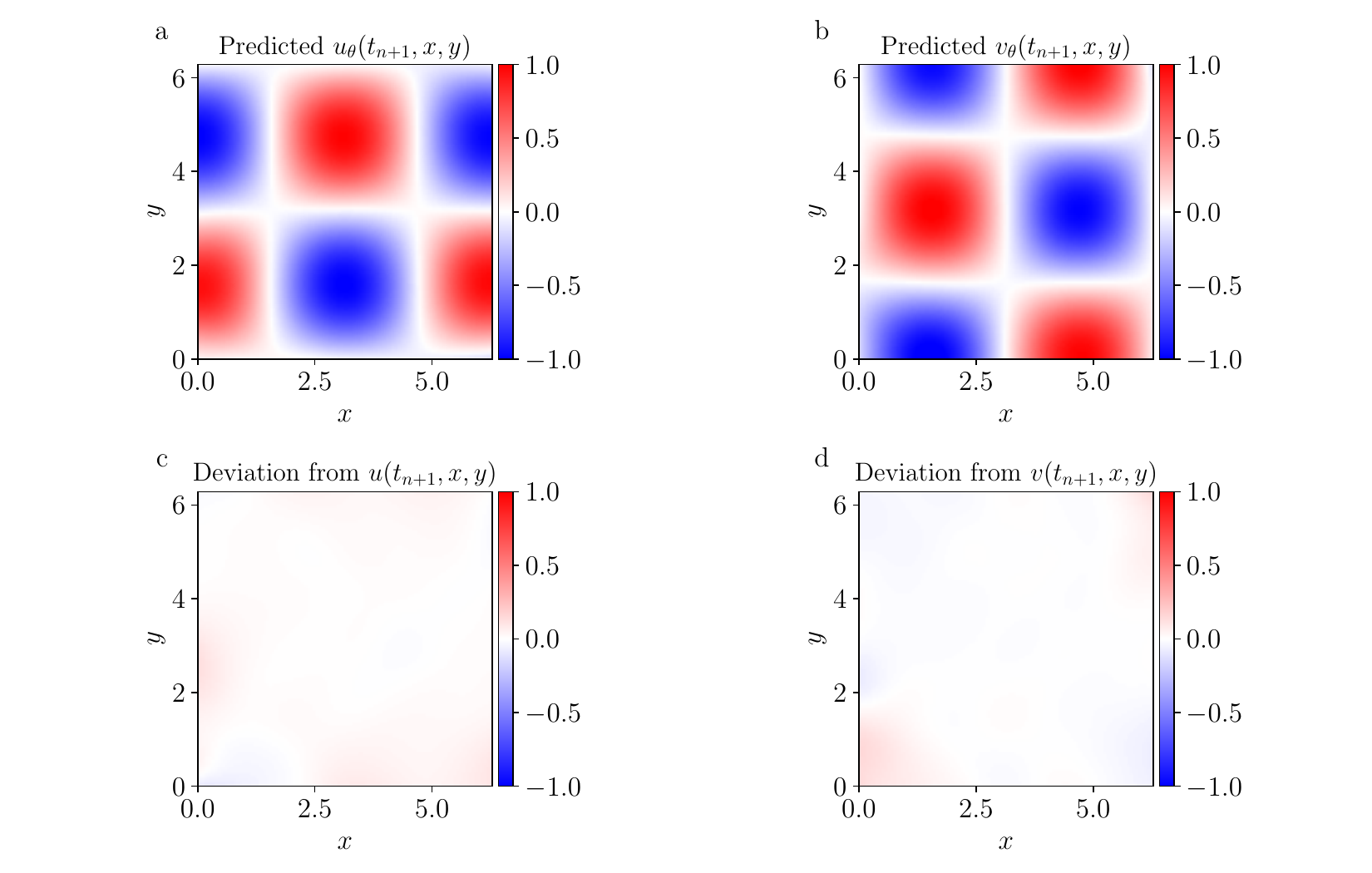}
   \caption[short]{Fit to the Taylor-Green vortex data with $\nu = 1$, $\rho=2$ using a IRK-PINN of
      order $q=100$ and time step $\Delta t = 1$. The training was carried out for 20000 epochs with
      ADAM and 20000 epochs with L-BFGS-B, using an MLP with 10 hidden layers with 16 nodes each and
      $SiLU$ activation function, and a sampling of $N=300$ points. (a) and (b) The results of the
      trained neural network prediction for $ u(t_{n+1},x)$ and $v(t_{n+1},x)$, respectively. (c)
      and (d) Deviation from the analytical solution in~\eqref{eq:TaylorVortex}. Total $L^1$-error
      $=2.01\%$}
   \label{fig:TG_vortex}
\end{figure}
For our simulations, we fixed both the viscosity $\nu=1$ and the mass density $\rho=2$. In addition,
we also fed the pressure $p(t, x, y)$ to the PINN such that its objective is only to predict both
components of the velocity fields $u$ and $v$. The results of the simulation and the details of the
neural networked employed are shown in \autoref{fig:TG_vortex}. The IRK-PINN is able to learn the
solution for a large time step with high accuracy, even when both the input and output dimensions
are not one-dimensional.

\section{Data availability}
\label{app:data_availability}
The code and data used of all example simulations is available at
\url{https://gitlab.desy.de/CMI/CMI-public/runge-kutta-pinn}.


\end{document}